\documentclass[runningheads]{llncs}
\usepackage[T1]{fontenc}
\usepackage{graphicx,verbatim}
\usepackage{url,array}
\usepackage{float}
\usepackage{amsmath}
\usepackage[hidelinks,breaklinks=true]{hyperref}

\begin{document}

\title{Physics-Guided Synthetic High-Frequency Ultrasound Generation for Skin Layer Segmentation}
\titlerunning{Physics-Guided Synthetic HFUS Generation}

\author{Junkyung Ju\inst{1} \and
Kyungho Yoon\inst{2}\and Minwoo Shin\inst{1}\thanks{Corresponding author.}}
\authorrunning{J. Ju et al.}
\institute{Department of Software, Yonsei University (Mirae Campus), Wonju, 26493, Republic of Korea\and School of Mathematics and Computing, Yonsei University, Seoul, 03722, Republic of Korea\\
\email{mshin@yonsei.ac.kr}}

\maketitle

\begin{abstract}
High-frequency ultrasound (HFUS) enables noninvasive visualization of superficial skin structures, but automated skin-layer analysis is limited by the scarcity of densely annotated data. Existing real HFUS datasets commonly provide annotations for superficial targets such as the epidermis and subepidermal low-echogenic band (SLEB), while dense labels for deeper structures such as dermis, subcutaneous tissue, fascia, and muscle are rarely available. We propose a physics-guided synthetic HFUS generation framework for skin layer segmentation. The framework constructs multilayer acoustic skin phantoms, assigns layer-dependent acoustic properties, and uses k-Wave simulation to generate paired synthetic HFUS images, dense layer masks, and simulation metadata. To evaluate whether the generated data provide transferable supervision, we use the generated synthetic data for downstream segmentation pretraining and fine-tune the models on a public real HFUS dataset. Synthetic pretraining followed by real fine-tuning achieved real-domain performance comparable to real-only training and improved mean Dice/IoU in three of the four trainable architectures evaluated. These results suggest that physics-guided synthetic HFUS images contain transferable anatomical and textural cues for real-domain skin layer segmentation, although further reduction of the synthetic-real appearance gap is needed to enable greater gains. The code and data are available at \url{https://github.com/Finn-02/synthetic-hfus-skin-layer-segmentation}.

\end{abstract}

\section{Introduction}
High-frequency ultrasound (HFUS) is widely used in dermatologic imaging because it provides high-resolution visualization of superficial tissue structures~\cite{barcaui2015skin,levy2021hfus}. In skin analysis, localizing the epidermis, SLEB, dermis, subcutaneous tissue, fascia, and deeper tissue regions is important for measuring thickness, monitoring structural changes, and supporting future image-guided applications. However, automated HFUS skin analysis remains difficult because images contain speckle noise, weak layer contrast, attenuation, probe-dependent appearance variation, and ambiguous tissue boundaries~\cite{czajkowska2021skinlayers,czajkowska2022dataset,loizou2008speckle}.

A major bottleneck is the scarcity of densely annotated HFUS skin datasets. Existing public resources mainly provide annotations for the epidermis and SLEB~\cite{czajkowska2021skinlayers,czajkowska2022epidermis,mendeley2021skinlayers}, whereas dense labels for dermis, subcutaneous tissue, fascia, and muscle are rarely available. Thus, supervised HFUS segmentation studies are often restricted to a narrow subset of skin anatomy.

Synthetic data generation can alleviate this limitation because simulation-based data can provide exact masks and generation metadata. Prior work has highlighted simulation and synthesis for medical image analysis under limited annotation settings~\cite{frangi2018sashimi}, and recent synthetic ultrasound studies show that generated images can support downstream learning~\cite{dahan2024csg,freiche2025ldm,stojanovski2023echo}. Nevertheless, skin-specific HFUS generation remains underexplored: existing skin HFUS studies mainly focus on real-image classification, quality assessment, or segmentation rather than physics-guided generation of densely labeled multilayer skin images. A useful synthetic dataset should therefore preserve the anatomical ordering of skin layers while allowing controlled variation in thickness, interface morphology, and tissue texture. This motivates a generator that is driven by acoustic layer maps rather than by image-level style transfer alone.

We propose a physics-guided synthetic HFUS generation framework for skin layer segmentation. The framework constructs multilayer acoustic skin phantoms, generates synthetic HFUS images using k-Wave~\cite{treeby2010kwave,treeby2012nonlinear}, and automatically exports dense 8-class masks and metadata. We then use downstream epidermis/SLEB segmentation to validate whether the generated data provide transferable supervision for real HFUS analysis.

Our contributions are as follows:
\begin{enumerate}
    \item We propose a physics-guided synthetic HFUS image generation framework based on multilayer skin acoustic phantoms and k-Wave simulation.
    \item We introduce controllable structural and appearance variations for layer-aware skin synthesis, producing dense labels for air, coupling medium, epidermis, SLEB, dermis, subcutaneous tissue, fascia, and muscle.
    \item We evaluate downstream utility through real-domain epidermis/SLEB segmentation, showing that synthetic pretraining followed by real fine-tuning achieves performance comparable to real-only training and improves mean Dice/IoU in three of the four trainable architectures.
\end{enumerate}

\section{Related Work}
\textbf{HFUS skin analysis and segmentation.} HFUS has been used to assess epidermal thickness, dermal echogenicity, SLEB-related changes, inflammatory dermatoses, skin aging, and aesthetic treatment effects~\cite{barcaui2015skin,levy2021hfus,nicolescu2023sleb,vergilio2021skinaging}. Learning-based HFUS studies have addressed image classification, quality assessment, epidermis segmentation, and epidermis/SLEB delineation~\cite{czajkowska2021skinlayers,czajkowska2022epidermis,czajkowska2021classification,czajkowska2022dataset}, with recent work extending automated segmentation to entry echo, SLEB, and dermis~\cite{slian2025attention}. These studies demonstrate the importance of automated HFUS skin analysis, but primarily operate on real images rather than physics-guided generation of densely labeled multilayer skin datasets.

\textbf{Synthetic ultrasound generation and simulation.} Data-driven ultrasound synthesis has been explored using diffusion and GAN-based models for echocardiography, breast ultrasound, and musculoskeletal ultrasound~\cite{dahan2024csg,freiche2025ldm,stojanovski2023echo,zama2023dcgan}. These methods show the value of synthetic ultrasound for downstream learning, but generally learn image distributions from data and do not explicitly construct multilayer skin acoustic maps with paired dense layer labels. Physics-based simulators provide a complementary route because the anatomical or acoustic model is known during image formation. Field II simulates ultrasound systems and transducer fields~\cite{jensen1996field,jensen1992pressure}, while k-Wave performs time-domain acoustic simulation in heterogeneous media~\cite{treeby2010kwave,treeby2012nonlinear} and includes B-mode simulation examples~\cite{kwaveBmode}. Despite the availability of these simulation tools, recent simulator reviews,
and learning-based acceleration methods for ultrasound acoustic simulation~\cite{shin2023multi,shin2024tfus,solano2025simulators}, physics-guided generation of densely labeled multilayer HFUS skin images for segmentation pretraining remains underexplored.

\section{Method}
\subsection{Overview of Physics-Guided Synthetic HFUS Generation}
\begin{figure}[t]
\centering
\small
\setlength{\tabcolsep}{2pt}
\begin{tabular}{@{}ccccc@{}}
\includegraphics[width=0.19\textwidth]{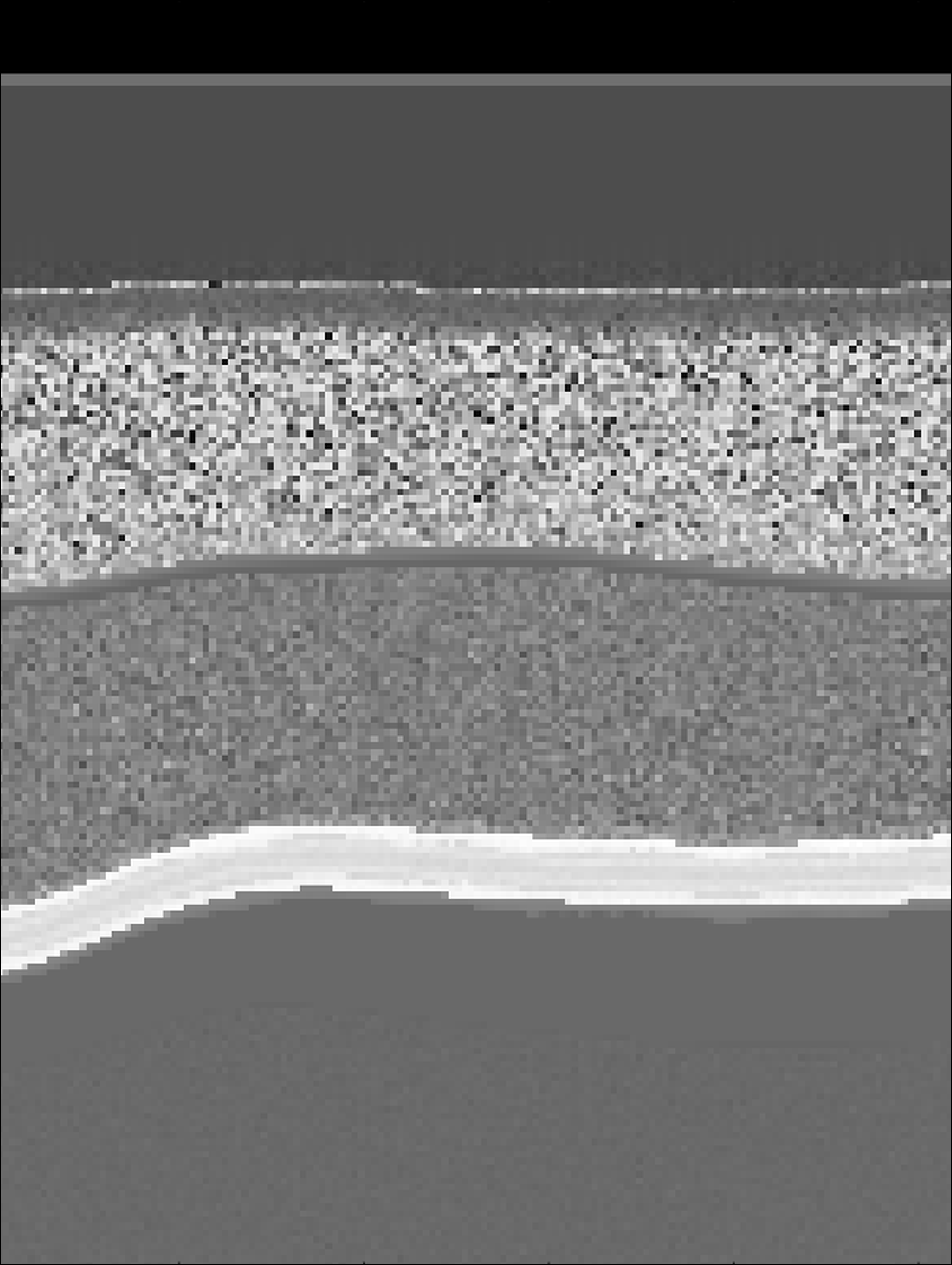} &
\includegraphics[width=0.19\textwidth]{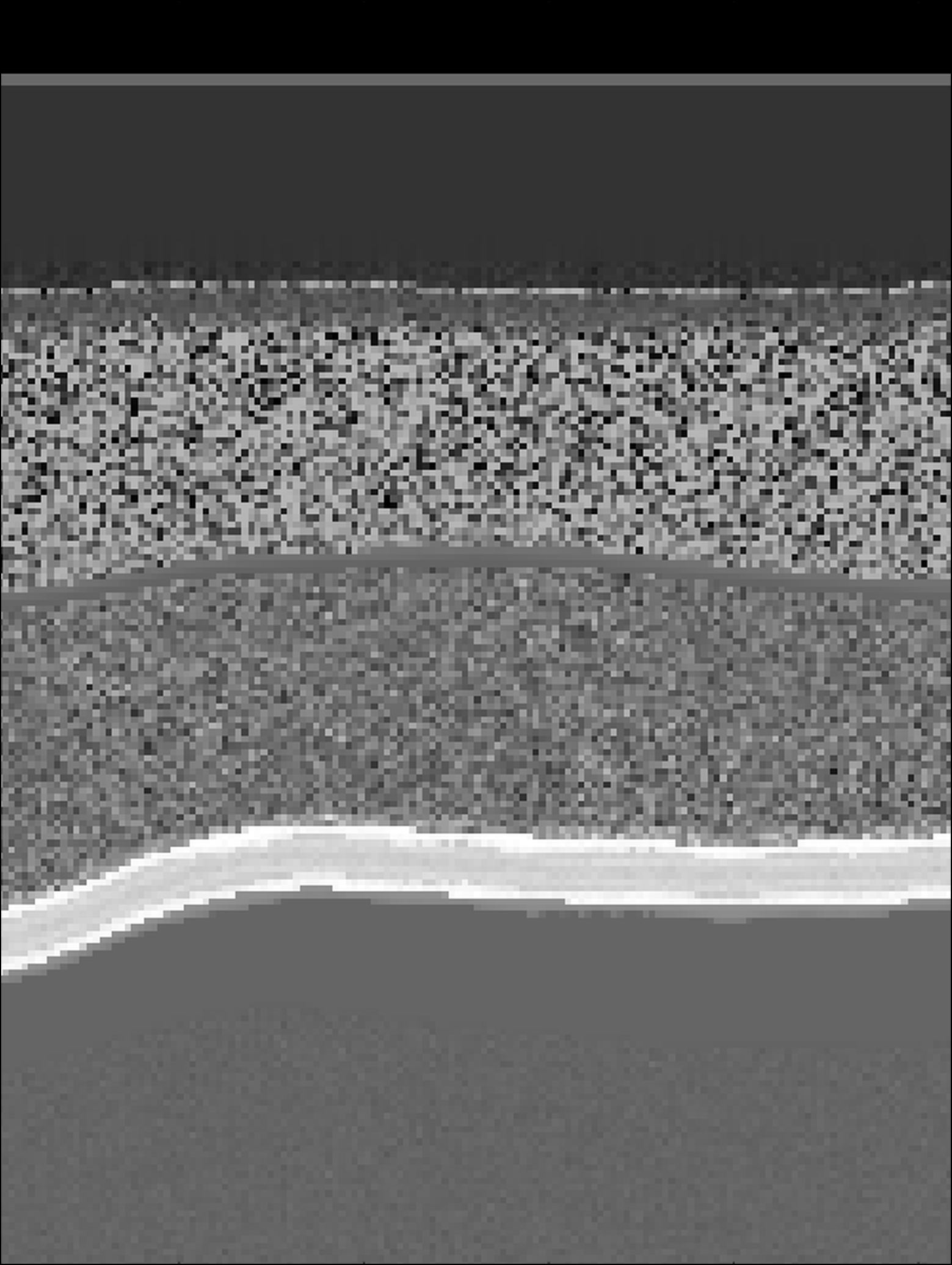} &
\includegraphics[width=0.19\textwidth]{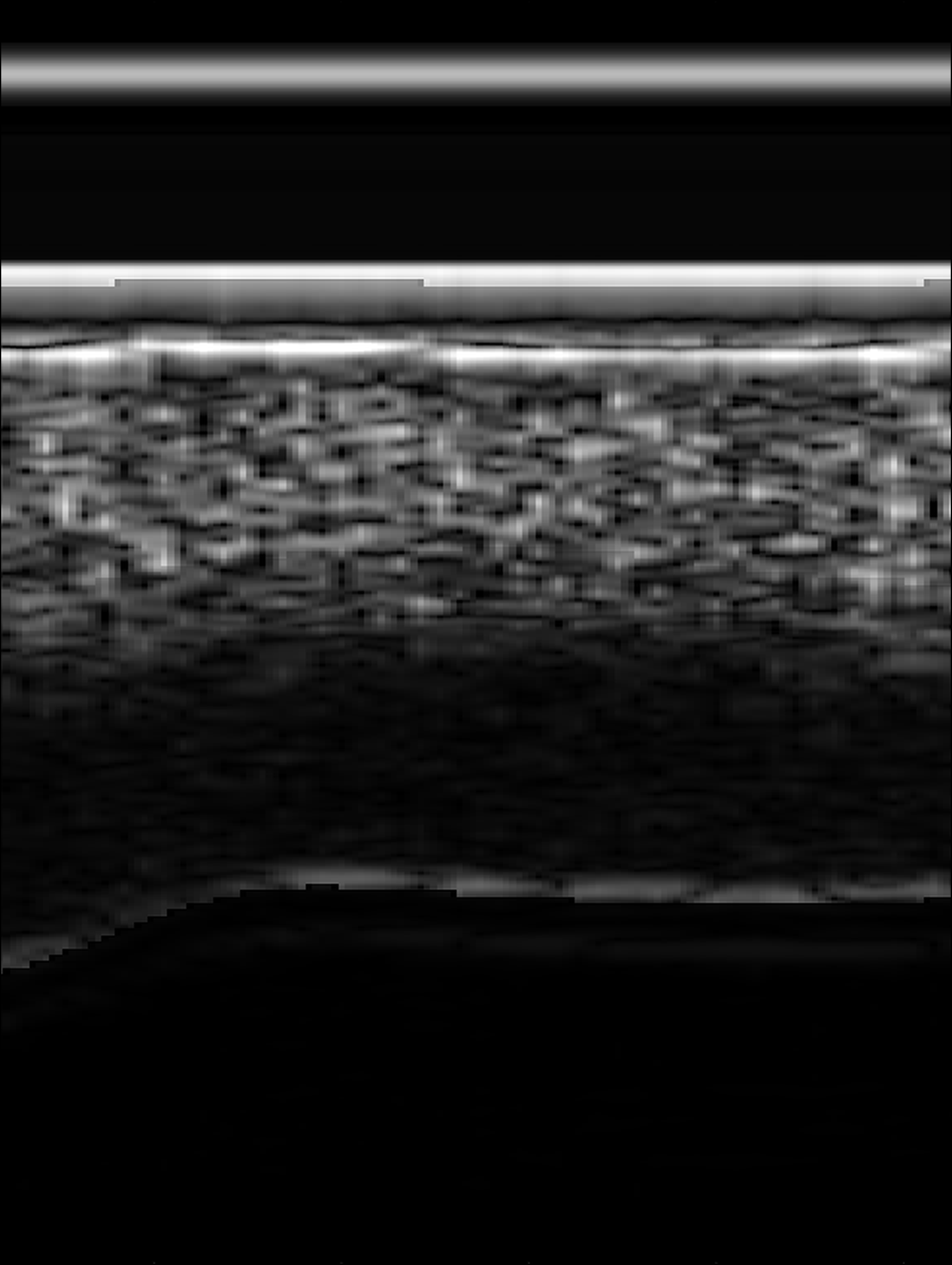} &
\includegraphics[width=0.19\textwidth]{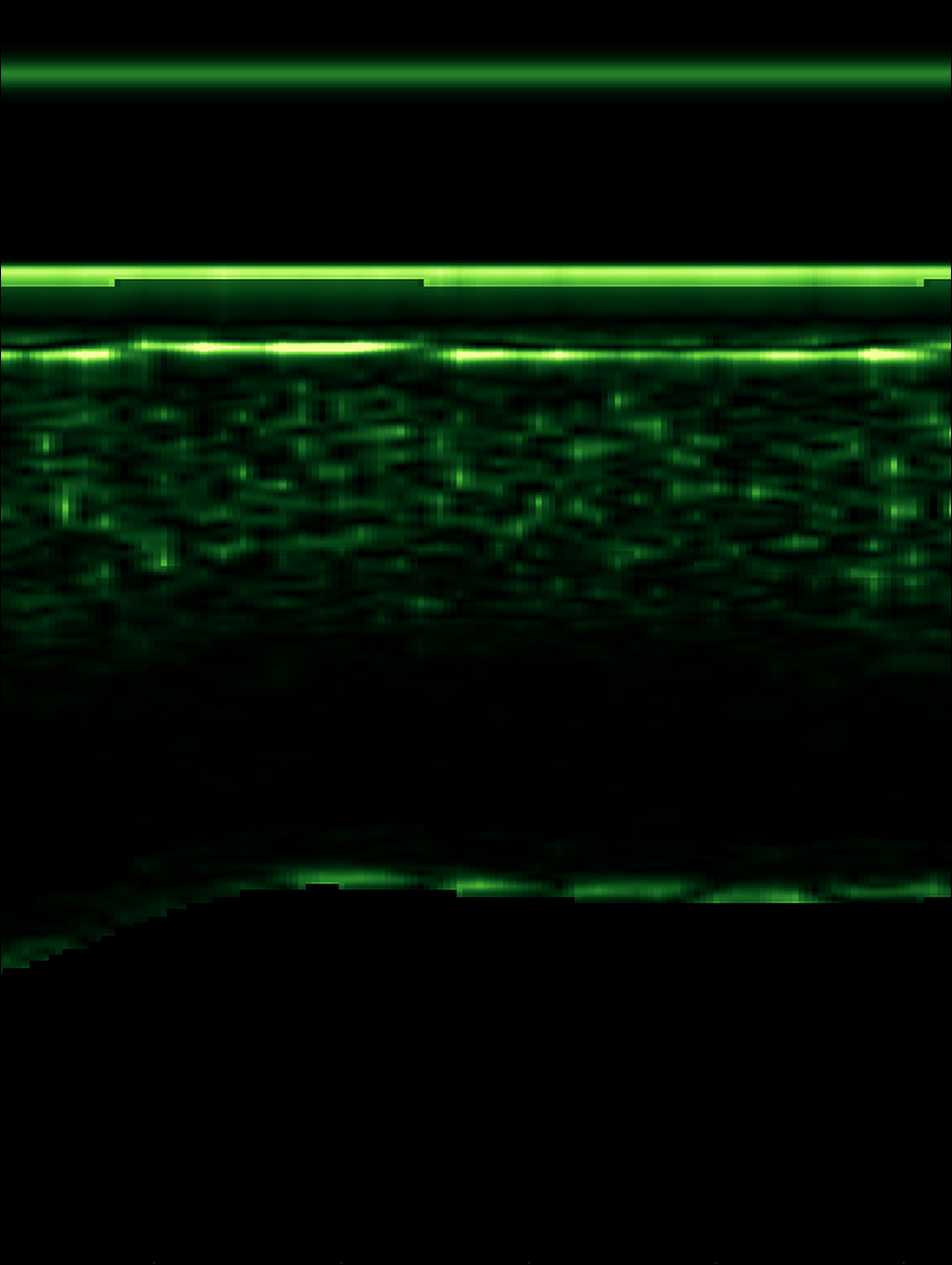} &
\includegraphics[width=0.19\textwidth]{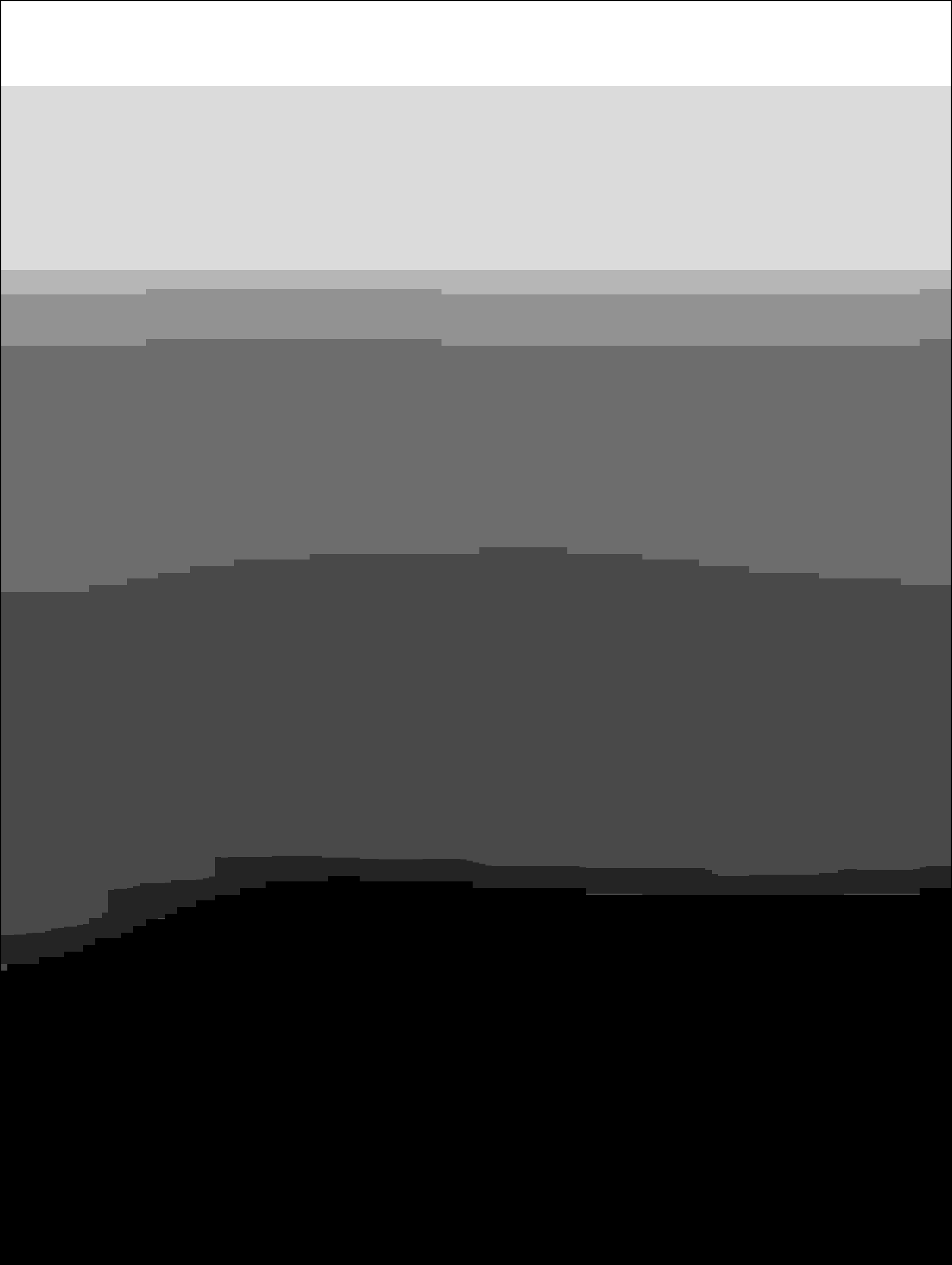}
\end{tabular}
\caption{Representative outputs of the proposed physics-guided synthetic HFUS generation framework. From left to right, the panels show the density map, sound-speed map, grayscale simulated ultrasound image, green-channel synthetic HFUS image, and dense layer mask. The acoustic property maps are simulated with k-Wave to generate ultrasound images, and the known anatomical layer map provides dense segmentation supervision.}
\label{fig:examples}
\end{figure}

Fig.~\ref{fig:examples} illustrates how the proposed simulator converts a multilayer acoustic skin phantom into paired synthetic HFUS images and dense segmentation masks. Starting from layer-dependent acoustic properties, the framework constructs density and sound-speed maps, performs k-Wave simulation, and exports both grayscale and green-channel synthetic HFUS images. Because the simulation is driven by known anatomical layer maps, the corresponding dense masks are obtained without manual annotation. This links acoustic phantom design, image generation, and pixel-level supervision within a single physics-guided pipeline. Each simulation output is stored together with its acoustic maps, raw class map, visual mask, overlay, and generation metadata. This organization makes the generation process traceable and allows downstream models to be trained on either the full multilayer label space or a task-specific label mapping.

\subsection{Multilayer Acoustic Skin Phantom}
The simulation pipeline uses k-Wave acoustic simulation to convert a multilayer acoustic skin phantom into synthetic HFUS images with raw class-ID masks~\cite{kwaveBmode,treeby2010kwave,treeby2012nonlinear}. The target label space contains eight classes: air, coupling medium, epidermis, SLEB, dermis, subcutaneous tissue, fascia, and muscle. Each class is assigned acoustic and structural properties through the simulator configuration. Table~\ref{tab:phantom_params} summarizes the principal material parameters used to initialize the phantom. SLEB is implemented as a low-scatter inserted band inside the upper skin region; its listed sound speed and density are target values used during SLEB material mixing.

\begin{table}
\centering
\small
\caption{Principal acoustic phantom parameters used in the synthetic HFUS generator. Sound-speed and density values were informed by reported acoustic properties of human skin, skin-mimicking phantoms, and soft tissues, while scatter levels are simulation controls for layer-dependent HFUS texture~\cite{chen2016skinphantom,lebertre2002skin,mast2000tissue,moran1995skin}.}
\label{tab:phantom_params}
\begin{tabular}{@{}lccc@{}}
\hline
Layer & Sound speed $c$ (m/s) & Density $\rho$ (kg/m$^3$) & Base scatter std. \\
\hline
Air / near-field medium & 1480 & 1000 & $2.0\times10^{-5}$ \\
Coupling medium & 1484 & 1002 & $2.0\times10^{-4}$ \\
Epidermis & 1628 & 1125 & $1.0\times10^{-2}$ \\
SLEB & 1518 & 1015 & low-scatter band \\
Dermis & 1578 & 1064 & $2.2\times10^{-2}$ \\
Subcutaneous tissue & 1556 & 1032 & $6.2\times10^{-3}$ \\
Fascia & 1512--1558 & 978--1036 & thin-sheet controlled \\
Muscle & 1560--1565 & 1040--1045 & $4.5$--$7.2\times10^{-3}$ \\
\hline
\end{tabular}
\end{table}

The simulator is designed to preserve thin superficial structures while extending the label space to deeper skin regions. The epidermis and SLEB are retained as separate classes. The dermis-to-subcutis transition is modeled as a gradual acoustic transition rather than a single artificial boundary, and fascia is represented as a thin physical layer at the subcutis--muscle interface. Muscle texture is modeled as a deeper anisotropic region rather than as a uniform background. This design allows the same synthetic image to support both the conventional epidermis/SLEB task and a broader multilayer segmentation task.

\subsection{Structural and Appearance Variation}
To avoid generating a single fixed skin geometry, the generator introduces three families of variation. First, layer-thickness variation changes the relative thickness of superficial and deeper layers across samples. Second, boundary-shape variation perturbs the morphology and smoothness of layer interfaces, allowing non-flat epidermal, SLEB, dermal, and subcutaneous boundaries. Third, tissue-heterogeneity variation modifies dermal scattering, subcutaneous hypoechoic texture, fascia visibility, and muscle texture. After generation, label-preserving image augmentation further increases appearance diversity while preserving the corresponding segmentation masks. We intentionally separate these controls: structural variation changes the simulated anatomy before image formation, whereas augmentation changes the exported image and mask pair after generation. This separation keeps anatomy-level diversity distinct from training-time appearance augmentation.

\subsection{k-Wave Simulation, Export, and Task Mapping}
The production workflow uses a 20 MHz simulation configuration with 96 scan lines and exports synthetic HFUS images at $2067\times1555$ pixels. This export size is intentionally matched to the publicly available real HFUS skin images used for comparison, which are also provided at $2067\times1555$ pixels~\cite{mendeley2021skinlayers}. The final field of view corresponds to a 7.0 mm axial depth.
To match the green display format of the publicly available real HFUS images used for
fine-tuning and evaluation, we use the fundamental green-channel rendering
as the segmentation input rather than the grayscale rendering.
The green-channel rendering is derived from the same simulated fundamental
image and does not alter the paired segmentation labels.
The raw integer mask independently preserves the class labels defined by
the common class map.

The generated dataset is organized as paired images, masks, visual masks, overlays, and metadata. In the current dataset, 200 original synthetic samples are produced from parent simulations and paired variants. For real-domain epidermis/SLEB segmentation, the synthetic 8-class masks are mapped into a 3-class target space: other, epidermis, and SLEB. Specifically, the epidermis class is retained as epidermis, the SLEB class is retained as SLEB, and all remaining synthetic classes are mapped to the other class. This mapping allows the synthetic data to be used as anatomical pretraining supervision while the final evaluation remains directly comparable to the public Mendeley HFUS epidermis/SLEB annotations.

\section{Experiments}
\subsection{Experiment 1: Generating the Synthetic Dataset}
The first experiment evaluates whether the proposed simulator can produce diverse synthetic HFUS skin images with paired dense labels. We generate synthetic samples by varying layer thickness, boundary morphology, and tissue heterogeneity, followed by label-preserving appearance augmentation. Because the class map is known before simulation, each synthetic image is paired with a dense mask and metadata without manual annotation. The generated samples therefore provide both image-level diversity and exact pixel-level supervision.

\subsection{Experiment 2: Downstream Segmentation Pretraining}
The second experiment evaluates whether the generated synthetic data provide transferable supervision for real HFUS segmentation. The real-domain evaluation uses the public Mendeley HFUS dataset, which provides real HFUS images and expert delineations for epidermis and SLEB~\cite{mendeley2021skinlayers}. Because real labels for dermis, subcutaneous tissue, fascia, and muscle are not available in this dataset, quantitative real-domain evaluation is restricted to epidermis and SLEB. The expanded multilayer label space is therefore demonstrated through the synthetic generation pipeline and dense mask output in Fig.~\ref{fig:examples}, rather than through real-domain multilayer accuracy.

The released SegUNet v0 model~\cite{czajkowska2021skinlayers} is included as a contextual reference and evaluated on the same extracted Mendeley ROI test set. The main comparison focuses on four trainable segmentation architectures: Fresh SegUNet, U-Net~\cite{ronneberger2015unet}, DeepLabV3+~\cite{chen2018deeplabv3plus}, and SegFormer~\cite{xie2021segformer}. For each trainable architecture, we compare real-only training with synthetic pretraining followed by real fine-tuning. This experiment evaluates the utility of the generated synthetic data, rather than proposing a new segmentation architecture.

All real-domain experiments use $64\times128$ ROI inputs with the same ROI extraction, split protocol, and Dice/IoU implementation. Performance is measured using class-wise Dice and intersection-over-union (IoU) for epidermis and SLEB, and the main table reports epidermis Dice, SLEB Dice, mean Dice, and mean IoU. All trainable models were optimized using stochastic gradient descent with momentum (SGDM), batch size 8, momentum 0.9, weight decay $5\times10^{-4}$, and early stopping with patience 40 and minimum delta $10^{-4}$. Real-only models were trained for up to 300 epochs, while synthetic-pretraining models used 100 synthetic epochs followed by up to 300 real fine-tuning epochs. Under the model-specific learning-rate policy, Fresh SegUNet used $10^{-3}$ for synthetic pretraining and $10^{-4}$ for real training/fine-tuning, whereas U-Net, DeepLabV3+, and SegFormer used $10^{-4}$ for synthetic pretraining and $10^{-5}$ for real training/fine-tuning.

\section{Results}
\subsection{Experiment 1: Synthetic Dataset Generation}
\begin{figure}[!htbp]
\centering
\small
\setlength{\tabcolsep}{2pt}
\begin{tabular}{@{}ccccc@{}}
\includegraphics[width=0.19\textwidth]{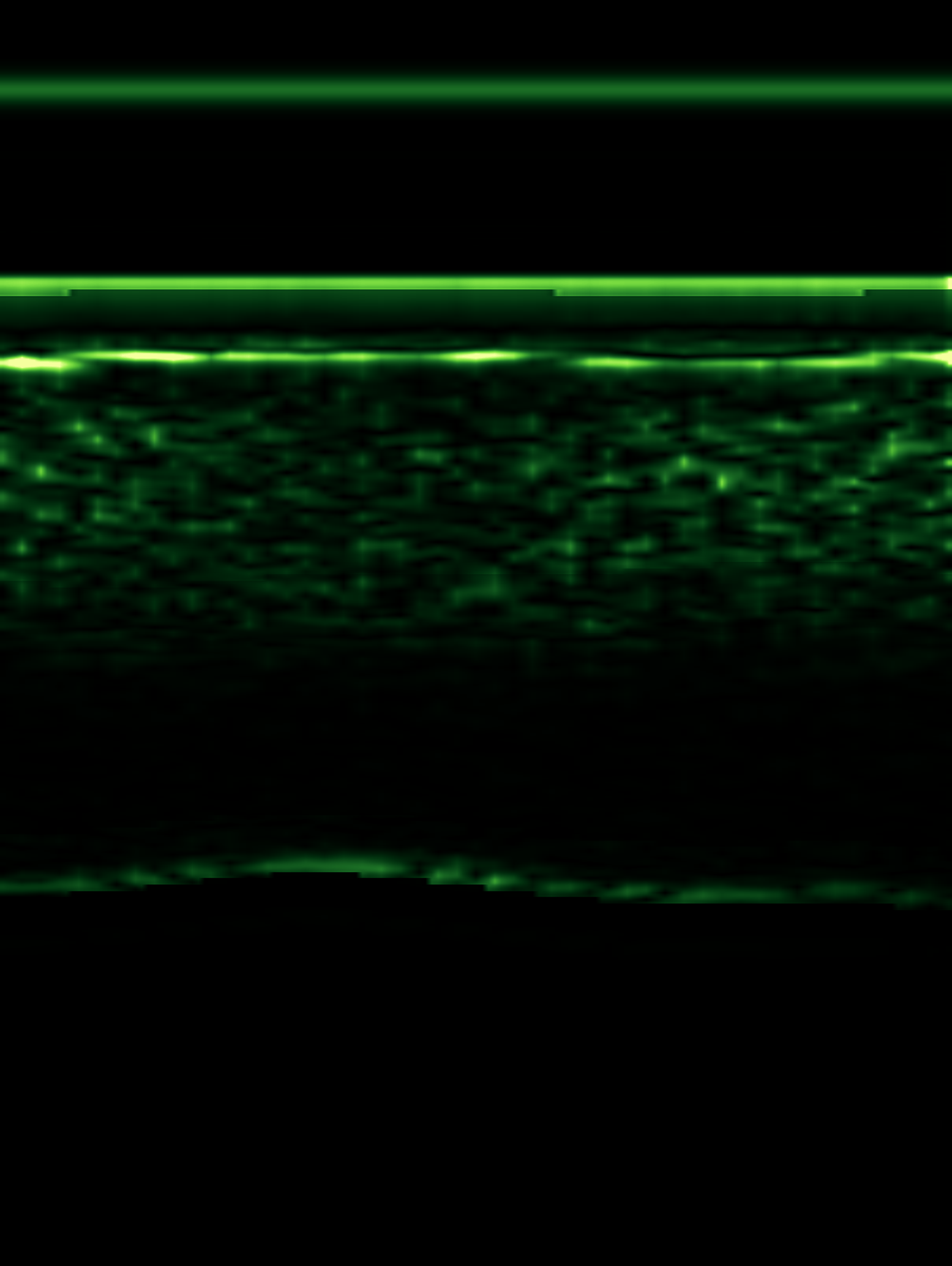} &
\includegraphics[width=0.19\textwidth]{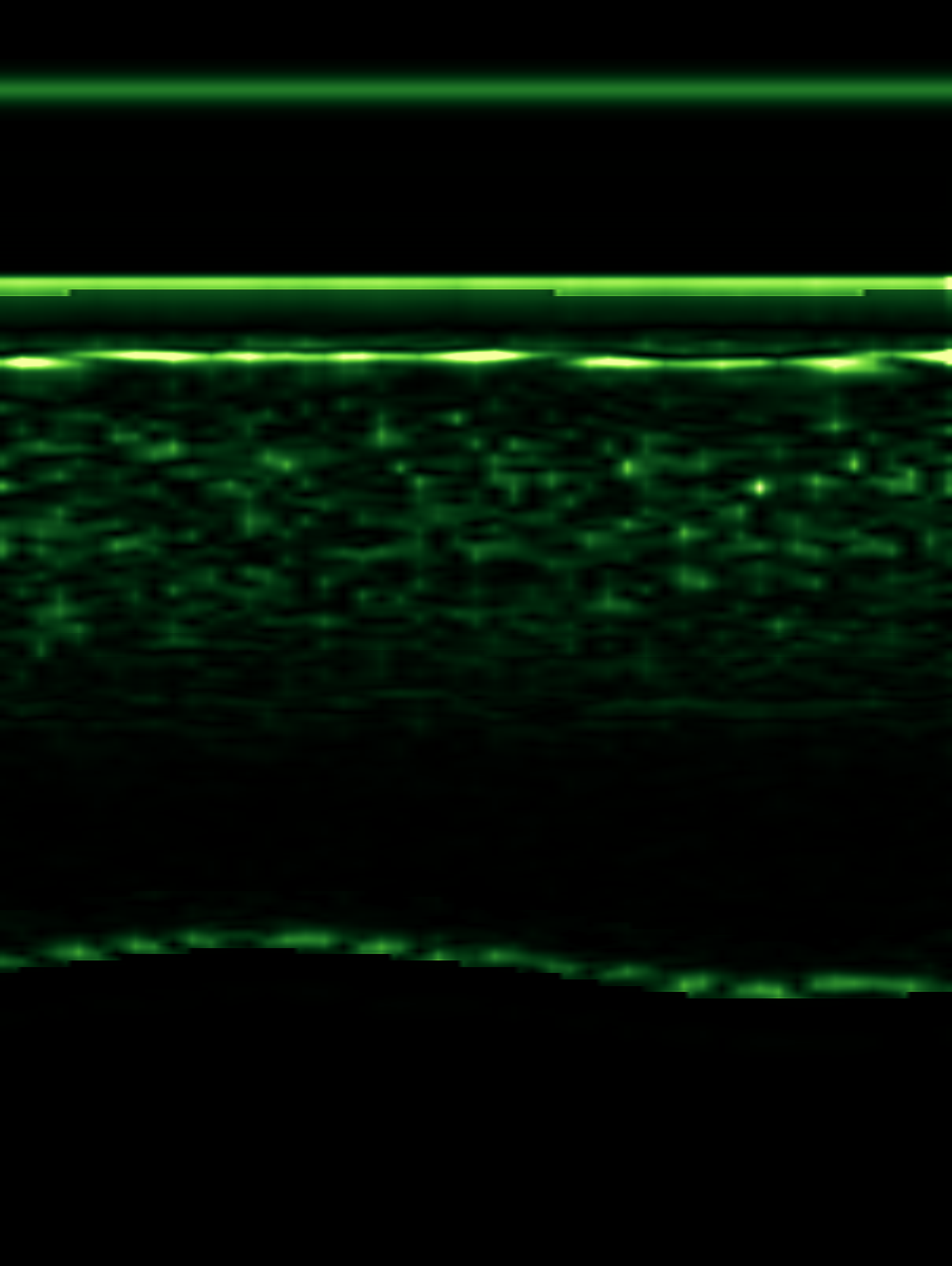} &
\includegraphics[width=0.19\textwidth]{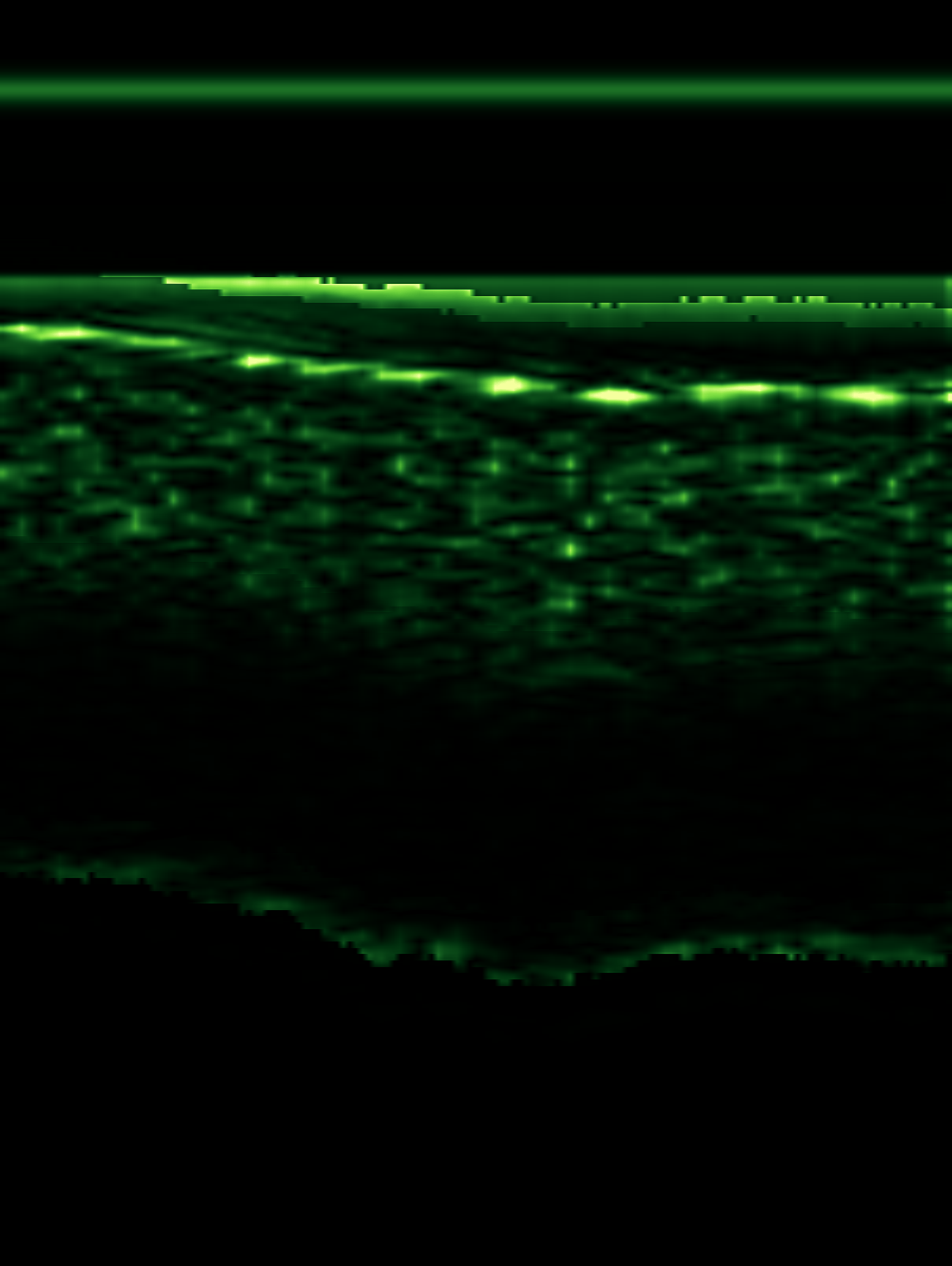} &
\includegraphics[width=0.19\textwidth]{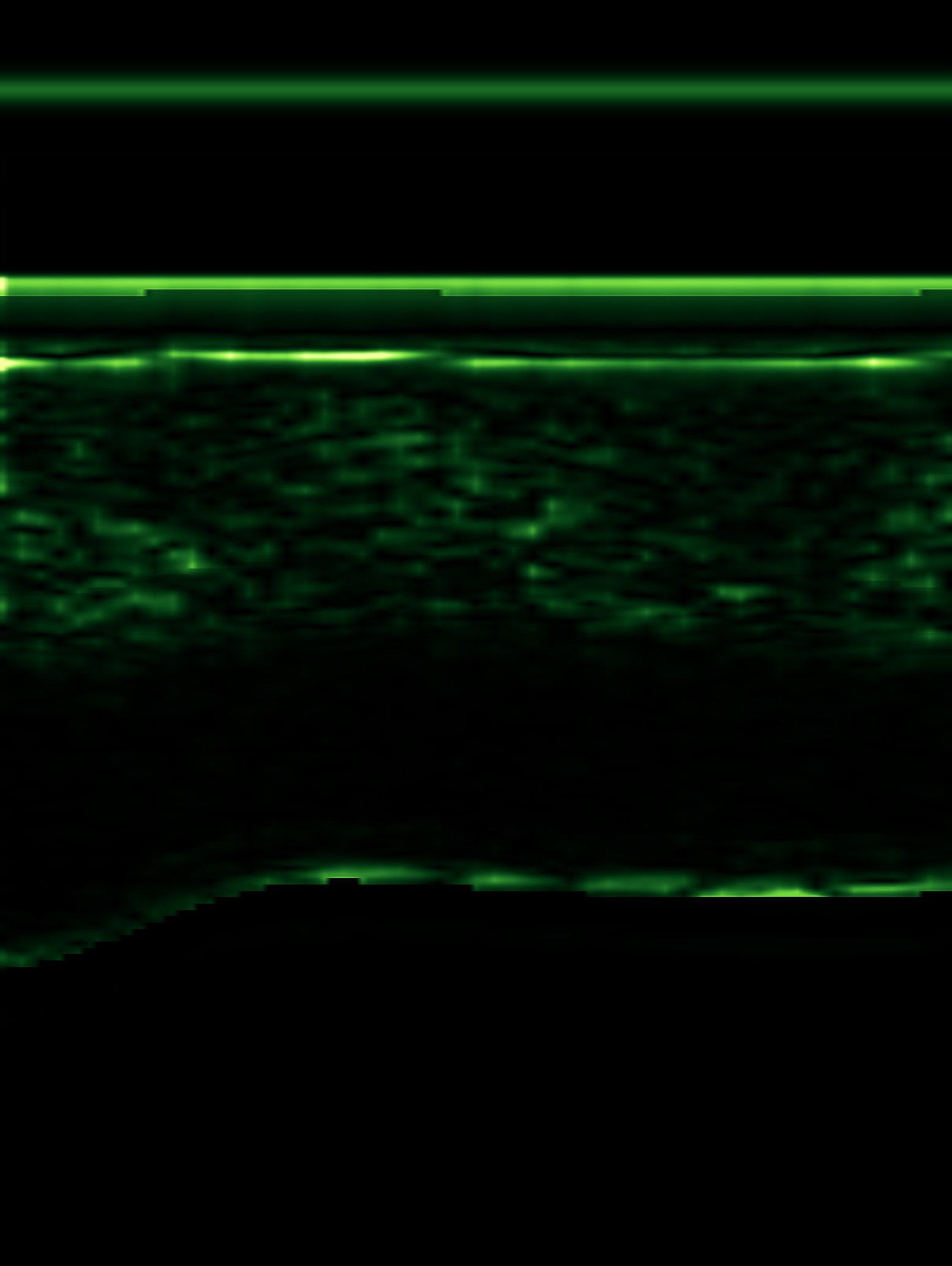} &
\includegraphics[width=0.19\textwidth]{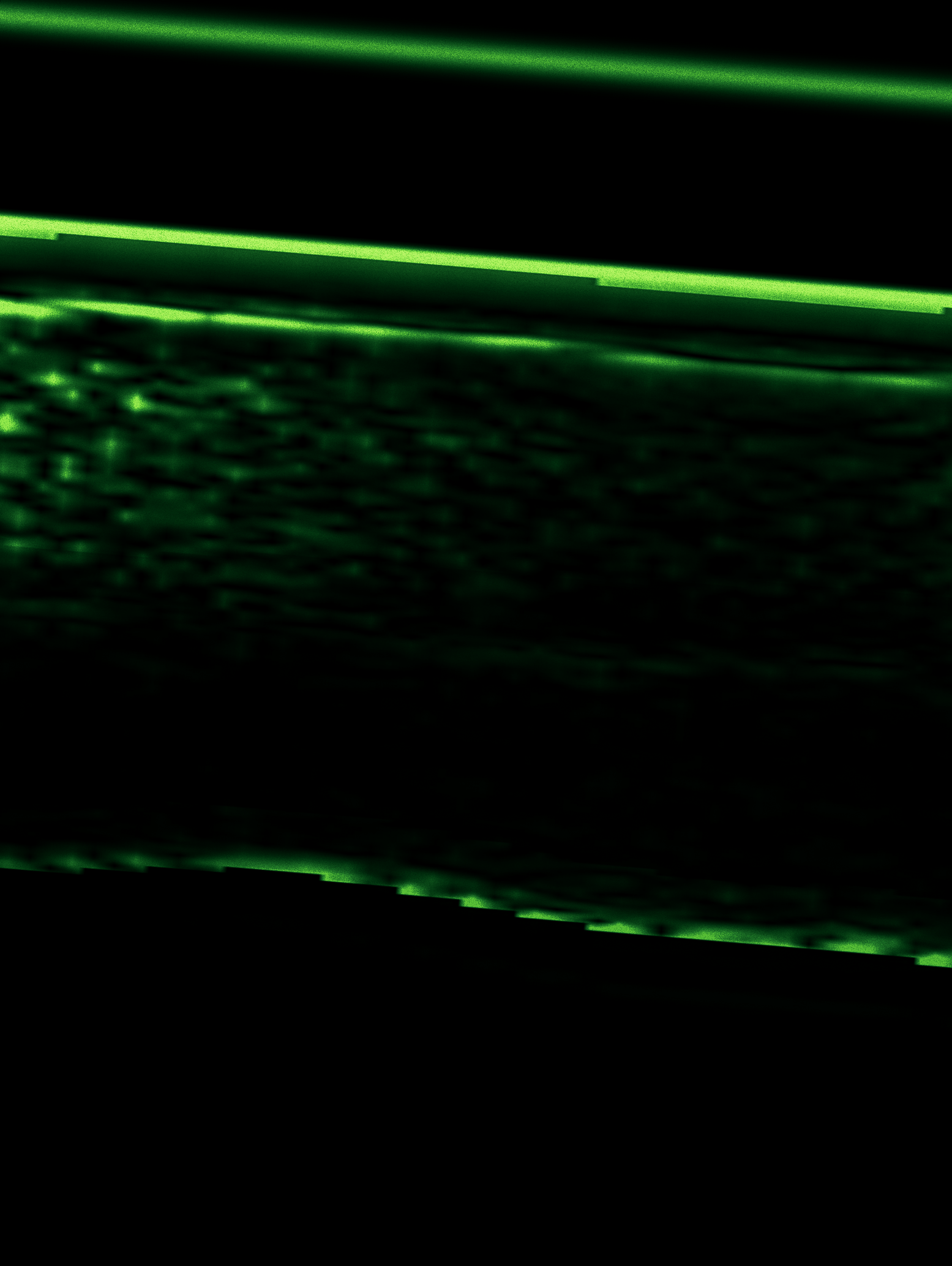}
\end{tabular}
\caption{Synthetic HFUS examples. From left to right: original synthetic image, thicker dermis variant, irregular boundary variant, stronger tissue heterogeneity, and rotation-based label-preserving augmentation.}
\label{fig:synthetic_variation}
\end{figure}

\begingroup
\emergencystretch=1em
Fig.~\ref{fig:synthetic_variation} shows representative synthetic HFUS images generated under structural and appearance variations. The examples illustrate baseline synthesis, thicker dermis, more irregular boundaries, stronger tissue heterogeneity, and label-preserving rotation augmentation. These samples are intended to show that the generator can vary both anatomical layout and visual texture while retaining exact pixel-level correspondence with the mask. In particular, the thickness and boundary examples demonstrate controllable structural diversity, whereas the heterogeneity and rotation examples demonstrate appearance diversity used for pretraining.
\par
\endgroup

\subsection{Experiment 2: Downstream Transferability Evaluation}
Table~\ref{tab:real_results} reports downstream epidermis/SLEB segmentation on the Mendeley test set. The purpose is to test whether synthetic HFUS images provide transferable supervision, not to establish a new state of the art in segmentation. Synthetic pretraining followed by real fine-tuning achieved performance comparable to real-only training and improved mean Dice/IoU in three of the four trainable architectures. The best mean Dice and mean IoU were obtained by Fresh SegUNet with synthetic pretraining and real fine-tuning, reaching 83.40\% and 71.79\%. Importantly, synthetic pretraining did not consistently degrade real-domain performance; even when the improvement was small, the pretrained models remained competitive after fine-tuning. Qualitatively, Fig.~\ref{fig:real_qual} shows that the SegUNet prediction after synthetic pretraining and real fine-tuning remains similar to the real-only result but yields a slightly more refined boundary in this example. The gains were modest, likely because synthetic images still differ from real scanner-acquired HFUS images in appearance.

\begin{table}[!htbp]
\centering
\footnotesize
\setlength{\tabcolsep}{2.2pt}
\caption{Downstream validation for real-domain epidermis/SLEB segmentation. Released SegUNet v0 is included as a contextual reference; the remaining rows follow the model-specific learning-rate policies described in the text. Mean Dice
and mean IoU are arithmetic averages over the epidermis and SLEB classes.
Bold indicates the best value for each metric. PT and FT denote pretraining and fine-tuning, respectively.}
\label{tab:real_results}
\begin{tabular}{@{}llcccc@{}}
\hline
Method & Training setting & Epi Dice & SLEB Dice & Mean Dice & Mean IoU \\
\hline
Released SegUNet & Released model & 78.14\% & 79.51\% & 78.82\% & 65.05\% \\
Fresh SegUNet & Real-only & 87.52\% & 77.75\% & 82.64\% & 70.71\% \\
Fresh SegUNet & Synth. PT + real FT & \textbf{87.91\%} & 78.90\% & \textbf{83.40\%} & \textbf{71.79\%} \\
U-Net & Real-only & 87.61\% & 76.64\% & 82.13\% & 70.04\% \\
U-Net & Synth. PT + real FT & 87.79\% & 78.39\% & 83.09\% & 71.35\% \\
DeepLabV3+ & Real-only & 83.00\% & 77.56\% & 80.28\% & 67.14\% \\
DeepLabV3+ & Synth. PT + real FT & 82.98\% & 77.06\% & 80.02\% & 66.80\% \\
SegFormer & Real-only & 84.67\% & \textbf{80.35\%} & 82.51\% & 70.28\% \\
SegFormer & Synth. PT + real FT & 85.42\% & 79.91\% & 82.66\% & 70.54\% \\
\hline
\end{tabular}
\end{table}

\begin{figure}[!htbp]
\centering
\small
\setlength{\tabcolsep}{2pt}
\begin{tabular}{@{}cccc@{}}
\includegraphics[width=0.24\textwidth]{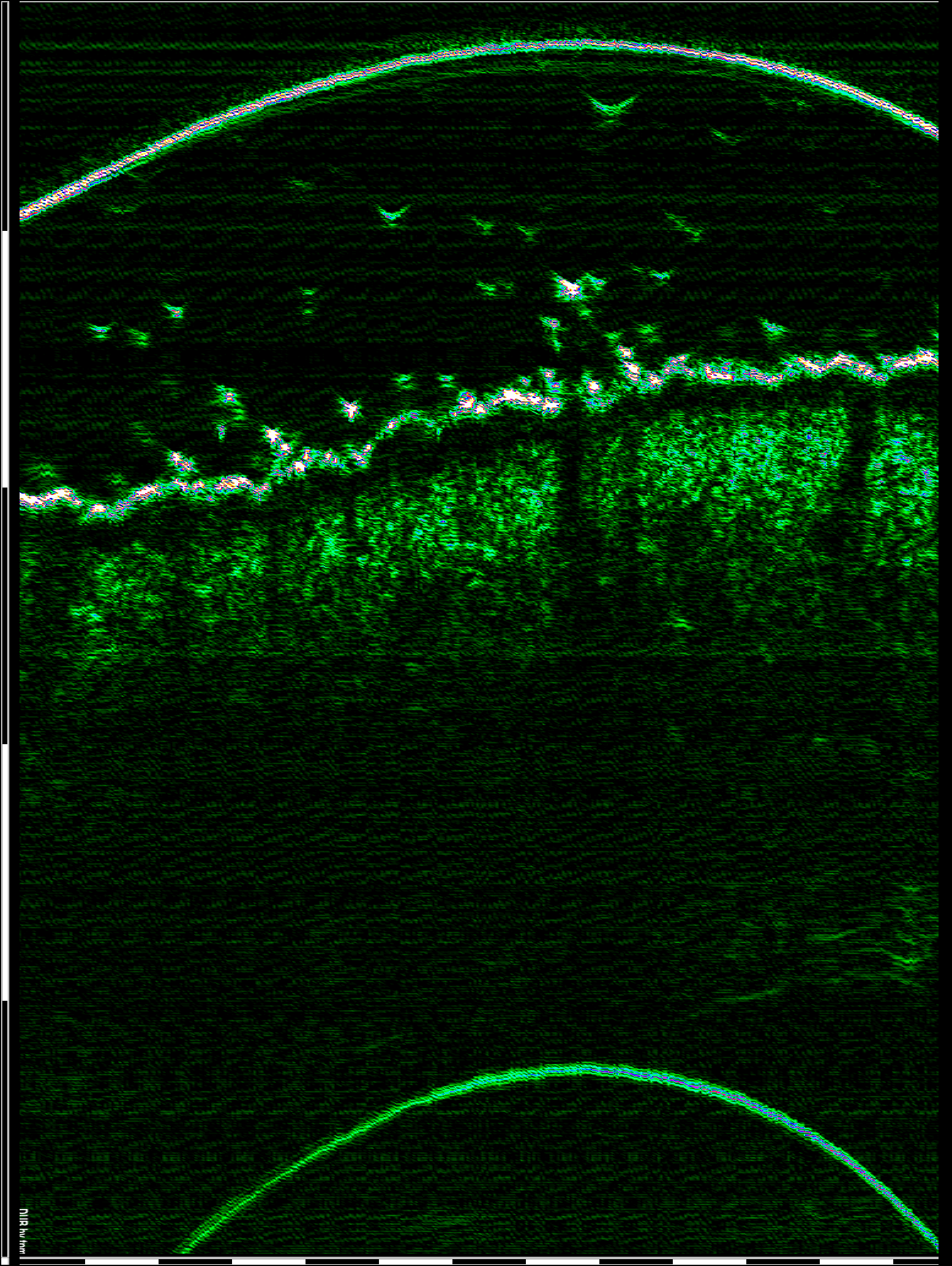} &
\includegraphics[width=0.24\textwidth]{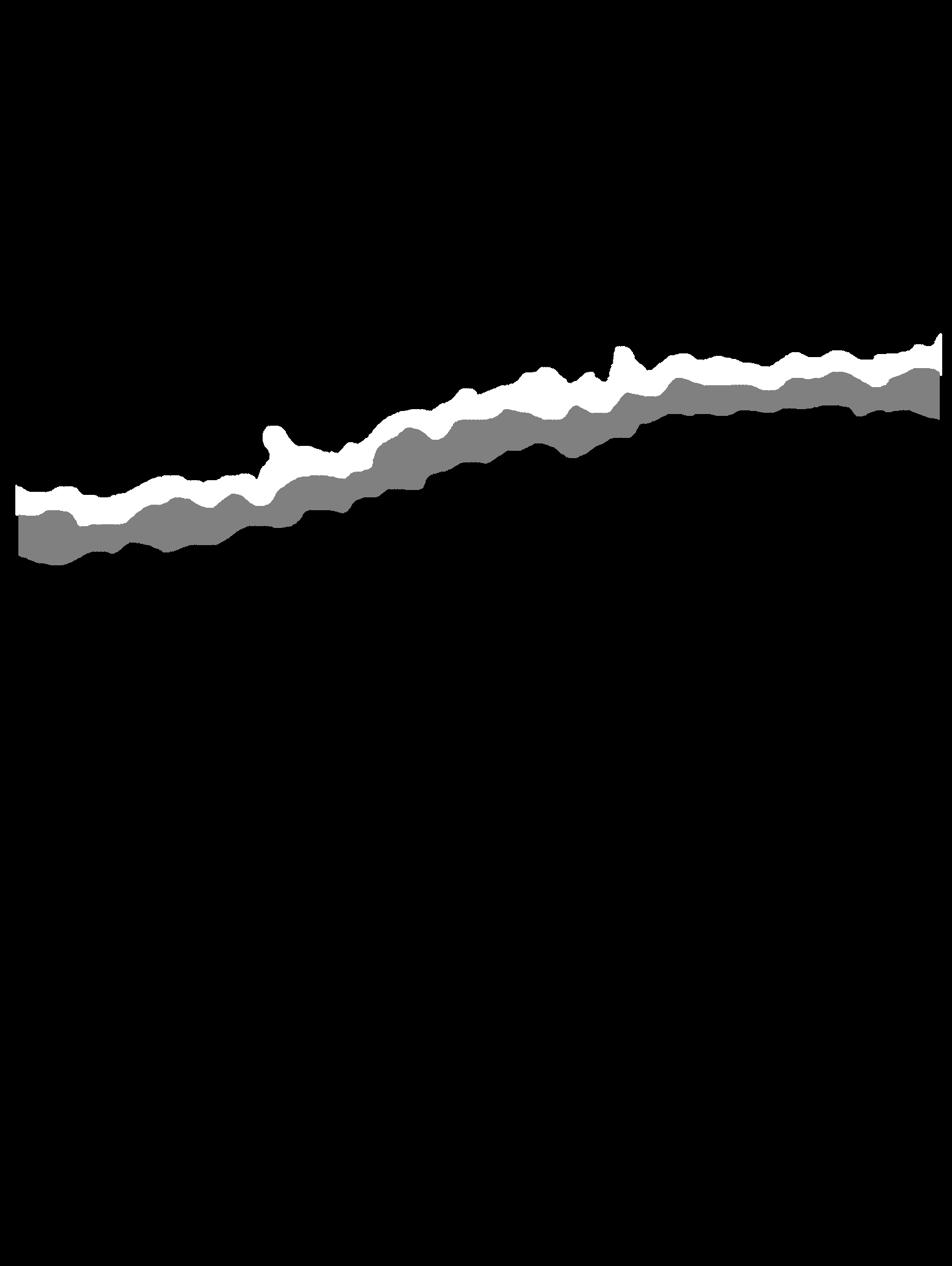} &
\includegraphics[width=0.24\textwidth]{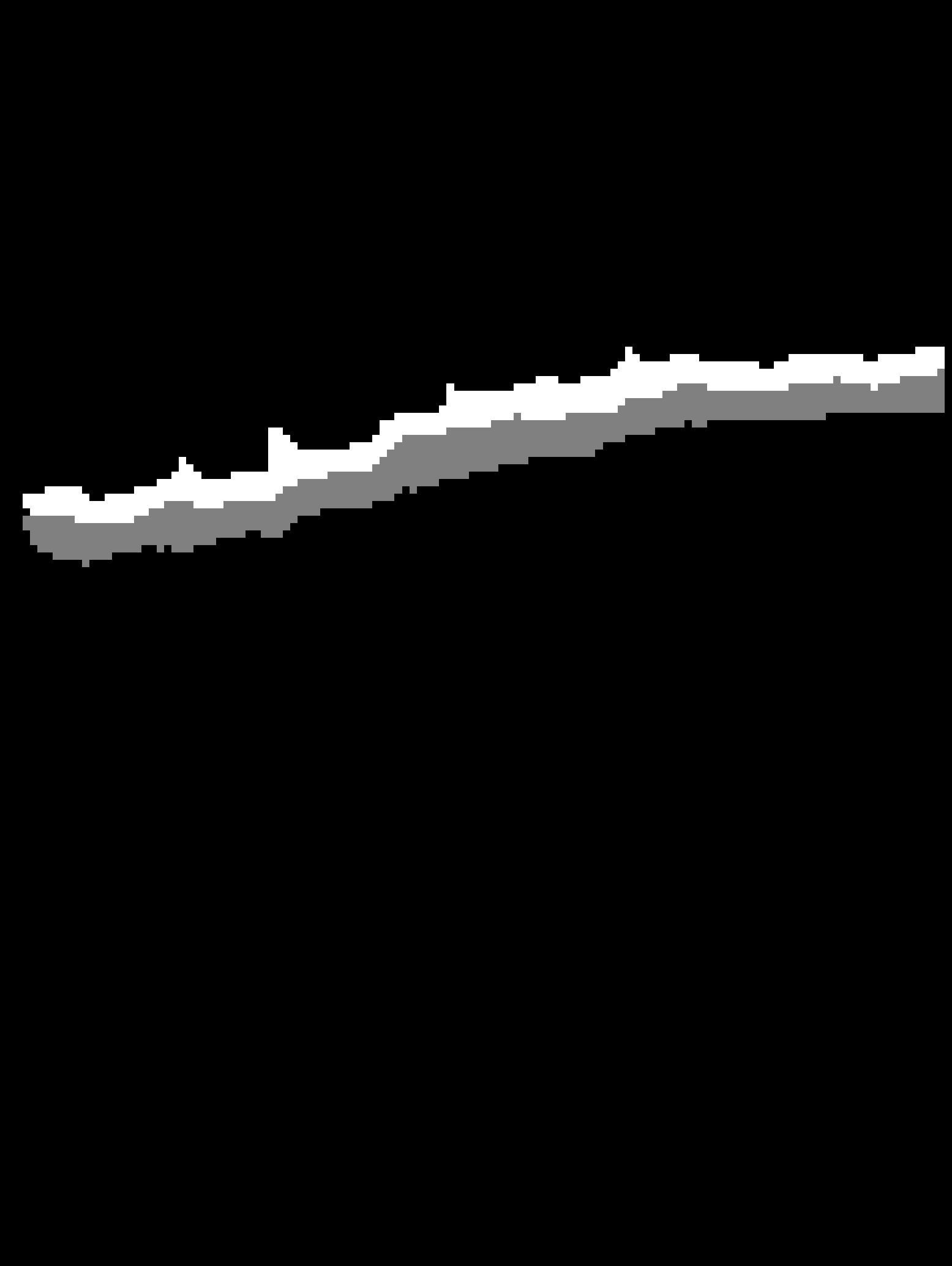} &
\includegraphics[width=0.24\textwidth]{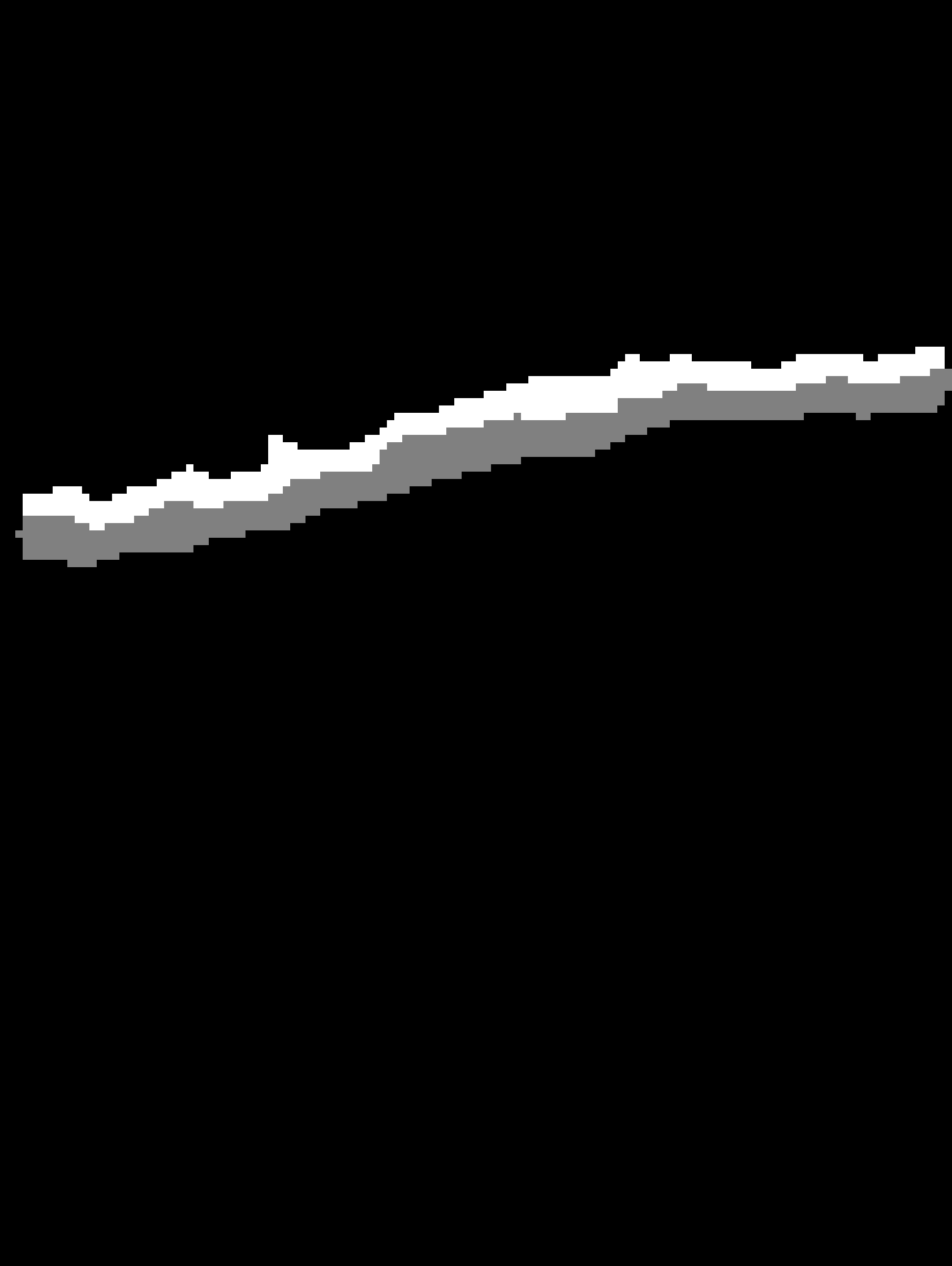} \\
\includegraphics[width=0.24\textwidth]{figures/fig03a_real_hfus.png} &
\includegraphics[width=0.24\textwidth]{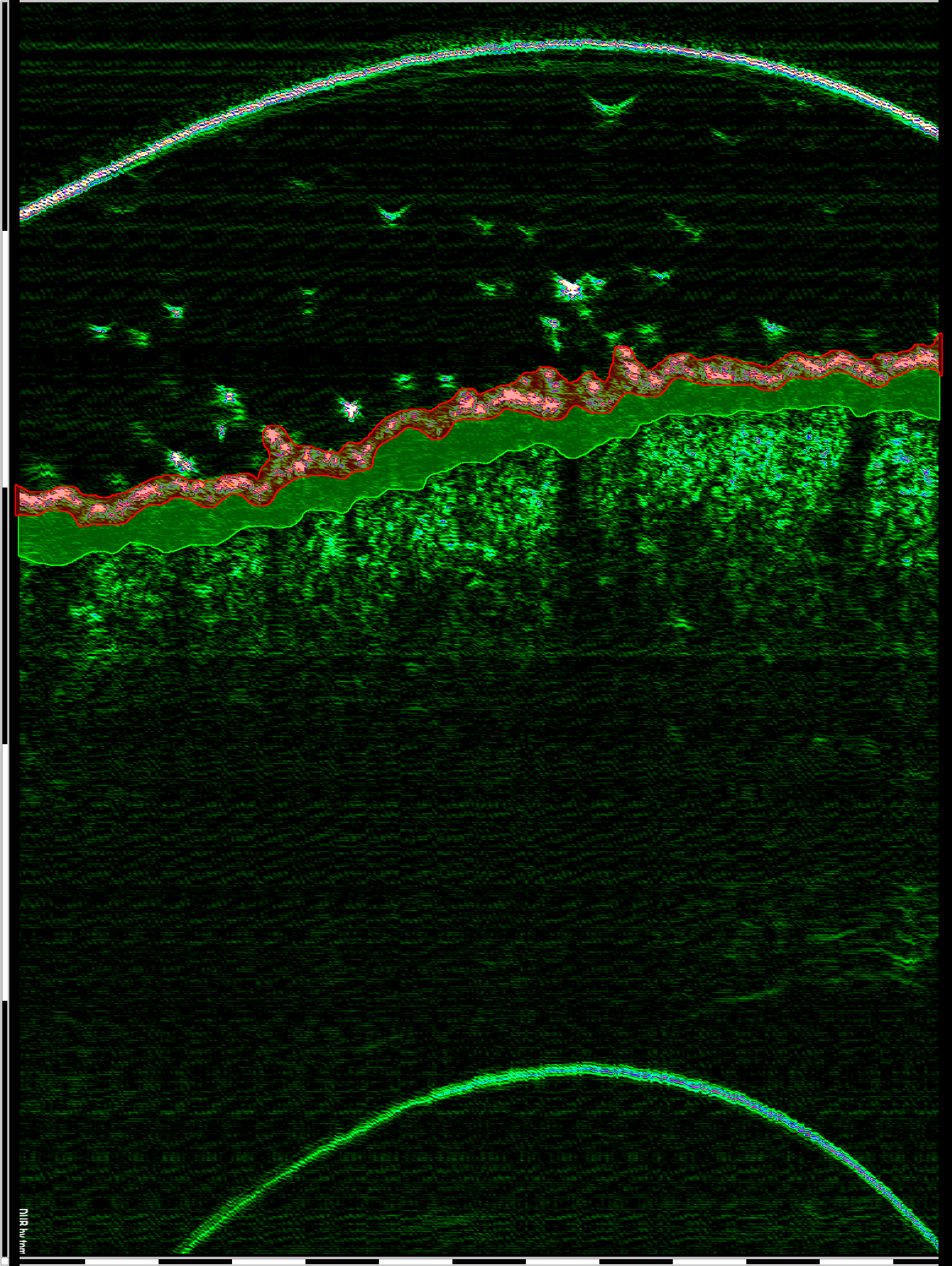} &
\includegraphics[width=0.24\textwidth]{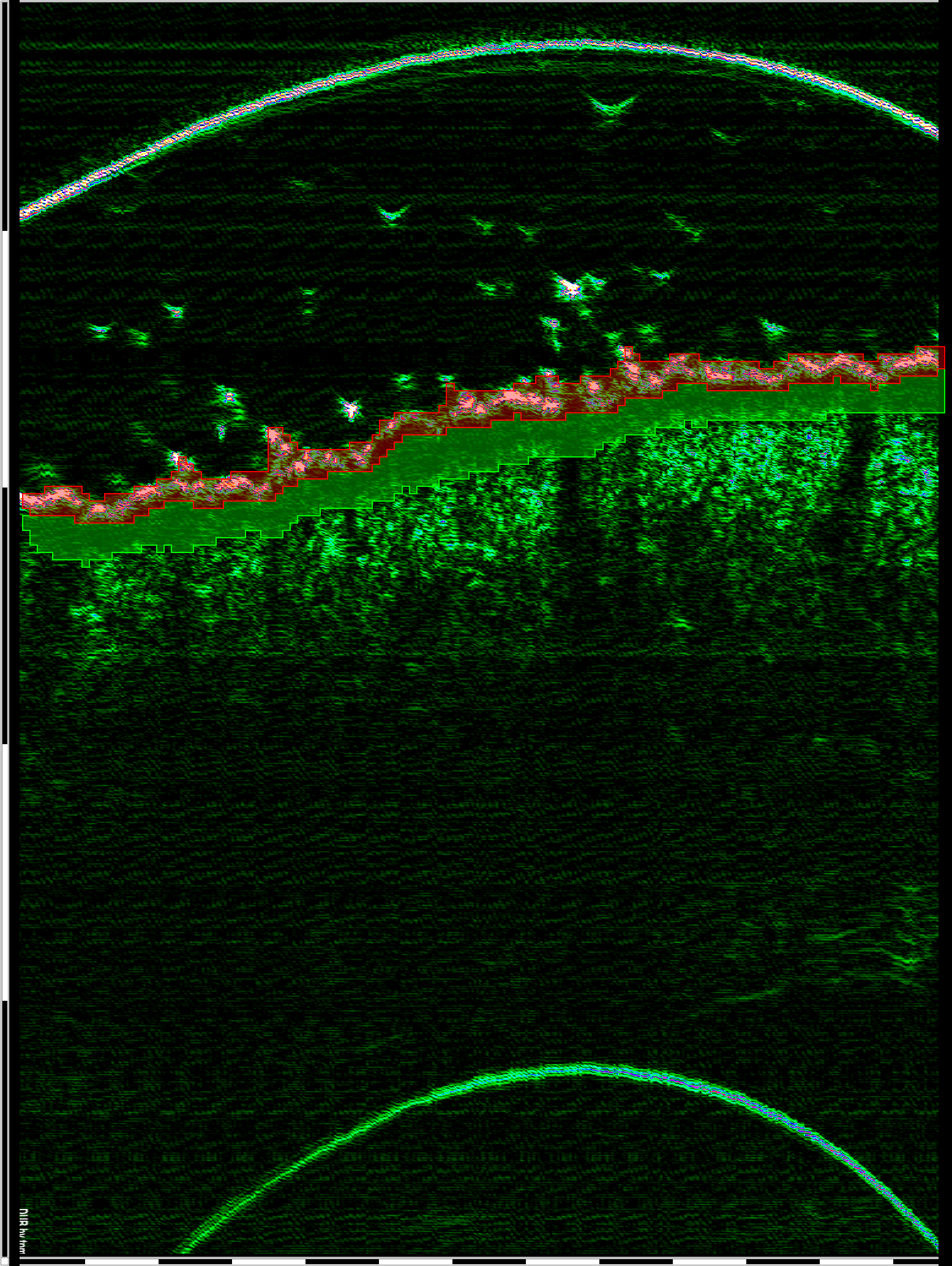} &
\includegraphics[width=0.24\textwidth]{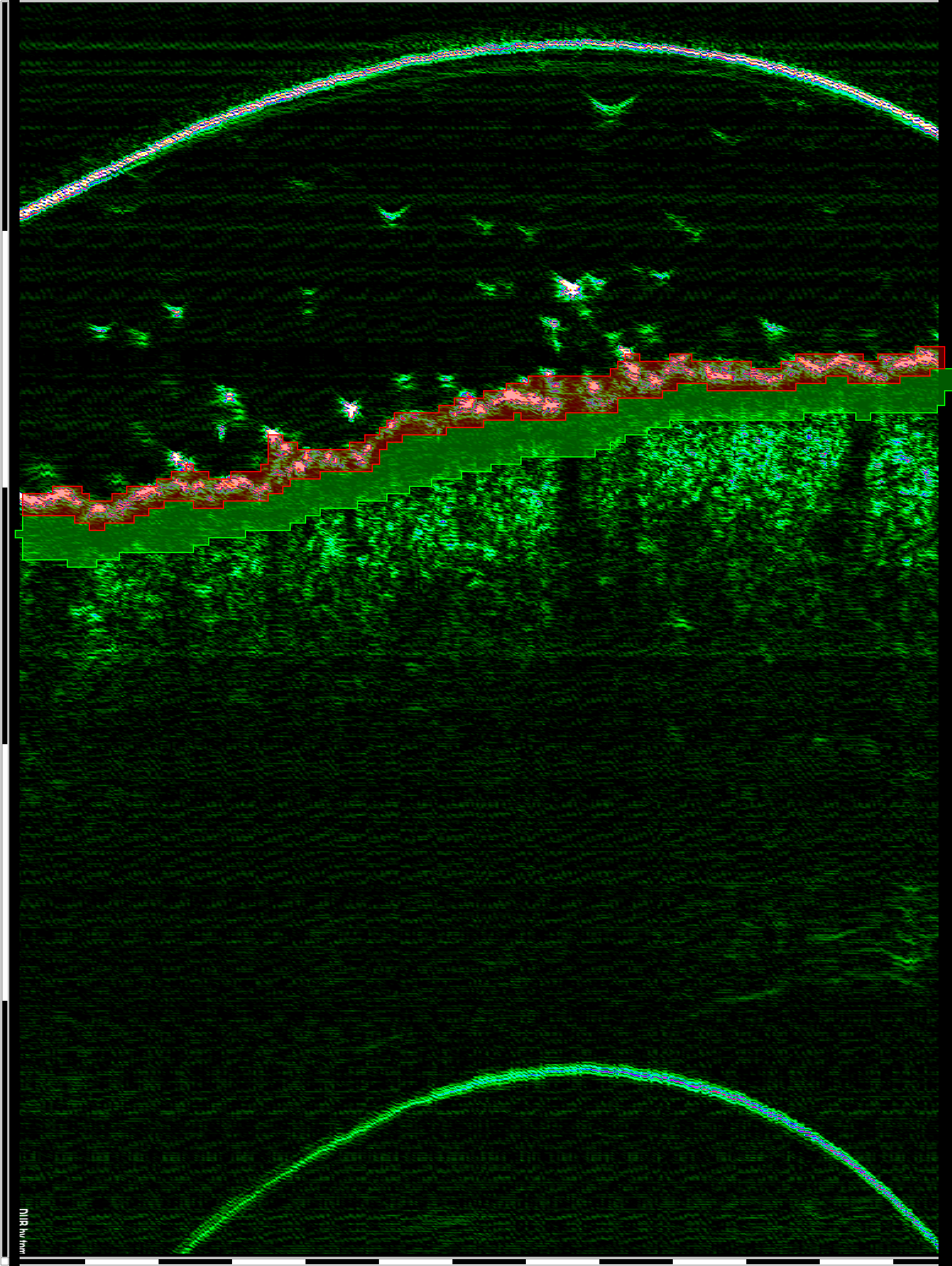}
\end{tabular}
\caption{Representative SegUNet result on a Mendeley case. Top row: original image, expert mask, real-only prediction, and synthetic-pretraining plus real-fine-tuning prediction. Bottom row: original image and corresponding overlays.}
\label{fig:real_qual}
\end{figure}

\section{Discussion}
The main contribution of this work is a physics-guided HFUS generation framework for multilayer skin imaging. In contrast to purely data-driven synthesis, it uses explicit acoustic layer maps to generate paired images, dense masks, and metadata.

The downstream experiment suggests that the generated images provide transferable anatomical and textural priors for real HFUS segmentation. After real fine-tuning, synthetic pretraining achieved performance comparable to real-only training and improved mean Dice/IoU in three of the four trainable architectures, indicating that the images are not merely visually plausible but encode layer-structured cues reusable by real-domain segmentation models.

The moderate improvement indicates that the current simulator captures transferable layer geometry and coarse tissue appearance, but does not yet fully match the HFUS statistics acquired by the scanner. This limitation is not only a matter of sample count: although learned generative models can sample efficiently after training, each physics-based sample requires costly wave-equation simulation. Increasing the number of samples generated by the same simulator may also preserve the same appearance bias. Differences in speckle distribution, attenuation behavior, boundary sharpness, and reflector artifacts may, therefore, explain the architecture-dependent gains.

A key advantage is the generation of dense multilayer labels beyond the real annotations available. Although real-domain evaluation is limited to epidermis and SLEB, the simulator also provides labels for dermis, subcutaneous tissue, fascia, and muscle. Reducing the remaining appearance gap, including overly strong reflector artifacts near the SLEB--dermis transition, is important for extending these initial transfer results toward more reliable multilayer HFUS segmentation.

\section{Conclusion}
We proposed a physics-guided synthetic HFUS generation framework for skin layer segmentation. The framework uses multilayer acoustic phantoms and k-Wave simulation to generate paired synthetic HFUS images, dense masks, and metadata. It produced structurally diverse synthetic samples and, in downstream segmentation, achieved real-domain performance comparable to real-only training while improving mean Dice/IoU in three of the four trainable architectures. These results suggest that the generated synthetic HFUS images provide transferable cues for real-domain skin layer segmentation. The framework therefore provides a practical basis for studying multilayer HFUS segmentation under controlled synthetic variation while retaining real-domain evaluation for the clinically annotated epidermis/SLEB task. We regard this as an initial validation of the usability of synthetic HFUS rather than a final solution to real-domain segmentation.

\begin{credits}
\subsubsection{\ackname}

This work was supported by the Korea Medical Device Development Foundation grant funded by the Korean government (the Ministry of Science and ICT, the Ministry of Trade, Industry and Energy, the Ministry of Health and Welfare, and the Ministry of Food and Drug Safety) (Grant No. RS-2026-25543484).

This research was also supported by the ANCHOR Program through the Gangwon ANCHOR Center, funded by the Ministry of Education (MOE) and Gangwon State (G.S.), Republic of Korea (Grant No. 2026-ANCHOR-10-006).

This research was further supported by the Ministry of Science and ICT (MSIT), Korea, under the National Program in Medical AI Semiconductor (Grant No. 2024-0-00096), supervised by the Institute of Information \& Communications Technology Planning \& Evaluation (IITP) in 2026.

\subsubsection{\discintname}
The authors have no competing interests to declare that are relevant to the content of this article.

\end{credits}

\bibliographystyle{splncs04}
\bibliography{Paper-0015}

\end{document}